\documentclass{article}
\usepackage{spconf,amsmath,graphicx}
\usepackage{multirow}
\usepackage{float}
\usepackage{xcolor}
\usepackage{mathrsfs}
\usepackage{amssymb}
\usepackage{setspace}
\title{token2vec: A Joint Self-Supervised Pre-training Framework Using Unpaired Speech and Text}
\name{Xianghu Yue$^1$, Junyi Ao$^2$, Xiaoxue Gao$^1$, Haizhou Li$^{2,1}$}
\address{
  $^1$Department of Electrical and Computer Engineering, National University of Singapore, Singapore \\
  $^2$School of Data Science, The Chinese University of Hong Kong, Shenzhen, China}
\begin{document}
\ninept
\maketitle
\begin{abstract}
Self-supervised pre-training has been successful in both text and speech processing. Speech and text offer different but complementary information. The question is whether we are able to perform a speech-text joint pre-training on unpaired speech and text.
In this paper, we take the idea of self-supervised pre-training one step further and propose token2vec, a novel joint pre-training framework for unpaired speech and text based on discrete representations of speech.
Firstly, due to the distinct characteristics between speech and text modalities, where speech is continuous while text is discrete, we first discretize speech into a sequence of discrete speech tokens to solve the modality mismatch problem.
Secondly, to solve the length mismatch problem, where the speech sequence is usually much longer than text sequence, we convert the words of text into phoneme sequences and randomly repeat each phoneme in the sequences.
Finally, we feed the discrete speech and text tokens into a modality-agnostic Transformer encoder and pre-train with token-level masking language modeling (tMLM).
Experiments show that token2vec is significantly superior to various speech-only pre-training baselines, with up to 17.7\% relative WER reduction. Token2vec model is also validated on a non-ASR task, i.e., spoken intent classification, and shows good transferability.
\end{abstract}
\vspace{-3mm}
\section{Introduction}
%The recent studies on representation learning of speech and text mimics the unsupervised language process of human language acquisition by first pre-training a model on a large amount of unlabeled speech or textual data, and then fine-tuning the model on a specific speech- or language-related downstream task.

Self-supervised representation learning of speech and text has evolved to a de-facto standard for speech and natural language processing (NLP), respectively. 
In speech, wav2vec2.0~\cite{w2v2}, HuBERT~\cite{Hubert} and their variants~\cite{data2vec, wavlm, pbert} have shown the powerful strength of self-supervised pre-training on speech recognition and many speech-related tasks~\cite{superb}. 
In NLP, GPT~\cite{gpt}, RoBERTa~\cite{roberta} and other variants~\cite{xlnet, spanbert} have enabled many language processing tasks to achieve the state-of-the-art performance (e.g., GLUE~\cite{glue}, SuperGLUE~\cite{superglue}, and XTREME~\cite{xtreme}).

Self-supervised learning models, like multilingual BERT~\cite{bert}, aim to be used universally, i.e. a single pre-trained encoder for many domains and languages. 
One big advantage of these unified models is the ability to leverage available unlabeled data across domains and languages~\cite{slam}. 
%GXX these unified models has an important advantage of their capability to leverage available unlabeled data across domains and languages~\cite{slam}. 
The availability of one language can boost the performance on other languages~\cite{bert}.
Speech and text are two different modalities but offer complementary information.
Therefore, jointly pre-training on unpaired speech and text is a natural next step to understanding both speech and text modalities.
Moreover, according to the language acquisition mechanism in humans, infants learn language by listening and reading simultaneously, which is a typical process that requires learning good joint representations of speech and text.
% GXX According to human language acquisition mechanism (citations), infants are able to learn language by listening and reading simultaneously, which resembles the typical process of learning meaningful joint representations of speech and text. This also motivates this work. 
% GXX ---- I feel you may need a paragraph summary and you may change the last second sentence to the last one. This paragraph is more like a motivation that would be helpful to inspire the proposed model from human perception perspective, so I suggest to move this paragraph/ part of paragraph to the second last paragraph of the introduction.

In terms of speech-text joint pre-training, there are two main branches. The first branch is to inject contextual knowledge of text into speech pre-training, since text pre-training can capture more high-level semantic information than that of speech.
SPLAT~\cite{spalt} and SAMU-XLSR~\cite{samu-xlsr} both pre-train a speech encoder to learn from a pre-trained text encoder using paired data. 
Another line of work usually employs two encoders for speech and text separately, based on which, a shared encoder is built to learn multi-modality knowledge. 
SLAM~\cite{slam} introduces two alignment losses, e.g. translation language modeling (TLM) and speech-text matching (STM) on paired data, to align speech and text. 
mSLAM~\cite{mslam} is a multilingual version of SLAM. 
Maestro~\cite{maestro} learns unified representations through sequence alignment, duration prediction, and modality matching based on RNN-T framework.
Most previous works ignore the inherent difference between these two modalities, which probably results in transferring interference and capacity limitations~\cite{slam}.

%Most previous work follow the same architecture with two separate speech/text encoders and a shared encoder, and try to learn the joint representations in a continuous latent space on limited paired data. These methods often suffer from capacity limitations and transfer interference between the two modalities~\cite{slam}. 

\begin{figure*}
    \centering
    \includegraphics[scale=1.5]{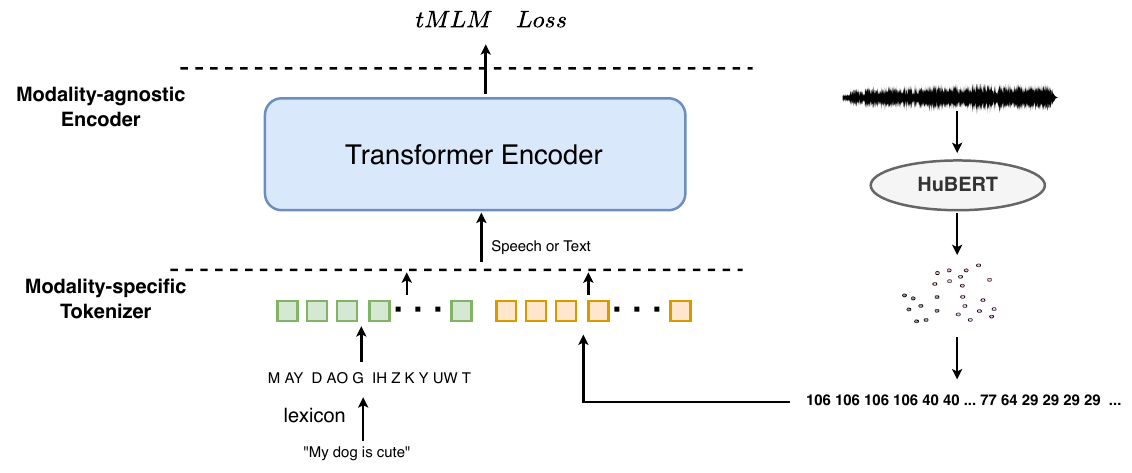}
    \caption{The framework of our proposed token2vec, consisting of two modules: (i) modality-specific tokenizers, which can take in speech and text inputs. The speech is tokenized into a sequence of discrete speech tokens, and the text is tokenized into a sequence of phonemes. (ii) modality-agnostic transformer encoder, which is shared by both speech and text. The token-level masked language modeling (tMLM) is designed for pre-training.}
    \label{fig:framework}
\vspace{-4mm}
\end{figure*}

In this paper, we propose token2vec, a novel joint pre-training framework, in which the learned speech representations can be further improved using unpaired text data. 
Specifically, we note that speech and text are different in nature, where speech frames are of continuous values, while text symbols are discrete tokens.
To solve this modality mismatch problem, we first propose to discretize continuous speech flow into a sequence of discrete speech tokens based on recent speech pre-trained models (e.g. HuBERT~\cite{Hubert}).
For text, we convert the word sequences into phoneme sequences according to a provided lexicon. 
By doing this, speech and text both become discrete sequences consisting of similar language units. 
Another obvious problem is that the speech sequence is usually much longer than text sequence, which will make the joint pre-training model difficult to converge. 
To solve this length mismatch problem, we up-sample the phoneme sequences by randomly repeating each phoneme according to the phoneme statistics, which is estimated on the subset of LibriSpeech dataset.
Finally, the discrete speech token sequences and the up-sampled phoneme sequences are fed to a modality-agnostic encoder iteratively to learn joint representations. 

We conduct massive experiments on the LibriSpeech~\cite{libri} dataset to validate the proposed token2vec. 
To the best of our knowledge, this is the first work to explore speech-text joint pre-training based on discrete speech tokens and text tokens.
Our method achieves better performance than various speech-only pre-training baselines. 
Moreover, we also evaluate the proposed token2vec on spoken intent classification task using Fluent Speech Command (FSC) dataset and show superior performance.

\vspace{-2mm}
\section{Related work}
We consider our work most related to Discrete BERT~\cite{discreteBERT}, in which the speech is quantized using vq-wav2vec~\cite{vq-w2v}, then representations learned on top by a BERT~\cite{bert} model. 
In Discrete BERT, the gumbel-softmax is used for vq-wav2vec quantization, and the model quantizes the speech into 13.5k unique codes.
vq-wav2vec~\cite{vq-w2v} builds a good vocabulary of the speech, which enables the effective speech representation learning in subsequent BERT training.
However, Discrete BERT only conducts the pre-training on discrete speech codes for speech representation learning, not leveraging the additional textual data for joint pre-training.
Based on Discrete BERT, our proposed token2vec takes the universality of self-supervised pre-training one step further, by unifying speech and text pre-training within one single model. 
Some recent works~\cite{spokenLM, audioLM, s2st} attempt to leverage discrete speech tokens to reconstruct speech, but they are mainly designed for audio generation and speech-to-speech translation tasks. 

In addition, another similar concurrent work is SpeechLM ~\cite{speechLM}, which aims to enhance speech representation with unpaired textual data. However, a small amount of labeled data must be used to train a tokenizer. 
In contrast, our method can achieve truly self-supervised speech-text joint pre-training without any labeled data.

\section{Method}
% GXX \section{Methodology}
In this section, we first illustrate the workflow of our proposed framework, based on which, we present each module in detail, including modality-specific tokenizer, modality-agnostic encoder and speech-text joint pre-training task, e.g., token-level masked language modeling.

\vspace{-2mm}
\subsection{Overall Framework}
Given unpaired speech and text, we aim to learn a unified representation of both modalities, which in turn improves downstream speech-related tasks.
Speech and text are two modalities with distinct characteristics, where speech is a continuous flow of signal, while text is composed of discrete tokens, and the length of the speech sequence is usually much longer than that of text sequence.
These two problems make speech representations difficult to benefit from additional textual data. 

In order to solve these problems, we propose to use modality-specific tokenizers to tokenize speech and text simultaneously, which will be described in detail later. 
After the tokenization processing, speech and text both are in the discrete space, which are defined as speech tokens and text tokens, respectively.
The speech tokens and text tokens are then fed to a modality-agnostic transformer encoder~\cite{transformer} to learn unified representations through token-level masking language modeling (tMLM).
Figure~\ref{fig:framework} illustrates the framework of the proposed token2vec.

\vspace{-2mm}
\subsection{Modality-speciﬁc Tokenizer}
Modality-specific tokenizers are employed to solve the modality mismatch problem and length mismatch problem as described above.
Specifically, inspired by Discrete BERT~\cite{discreteBERT}, we first tokenize the continuous speech into discrete speech tokens. 
Following HuBERT~\cite{Hubert}, we firstly extract the representation of 9-th layer of HuBERT, and train a k-means model with 500 centroids on these representations. 
Then, we take the index of the closest centroid to represent the speech. 
In this way, a continuous speech sequence is converted into a discrete speech token sequence.

For text, we firstly use a lexicon to convert word sequences into phoneme sequences, since the discretized speech units are closer to phonemes. 
Then we up-sample the phoneme sequences by repeating each phoneme several times to make sure that speech and text have similar lengths. 

Formally, a continuous speech sequence is processed into a sequence of tokens as follows:
% To summarize, a continuous speech sequence 
\begin{equation}
    S = [s_1 U, \dots, s_M U] + U^{pos}
\end{equation}
where $U$ is the speech token embedding matrix, $U^{pos}$ is the position embedding for speech. For text, a sequence of words is tokenized into a sequence of phonemes as following:
\begin{equation}
    T = [p_1 V, \dots, p_N V] + V^{pos}
\end{equation}
where $V$ is the phoneme embedding matrix, $V^{pos}$ is the position embedding for text.

\subsection{Modality-agnostic Encoder}
The Transformer encoder consists of stacked blocks including a multi-head self-attention (MSA) layer and a MLP layer. The MLP contains two fully connected layers with a GELU non-linearity. Layer normalization (LN) is applied before each MSA or MLP layer:
\begin{align*}
    & z^0 = S \ or \ T  \\
    & \hat{z}^l = MSA(LN(z^{l-1})) + z^{l-1}, \quad l=1, \dots, L \\
    & z^l = MLP(LN(\hat{z}^l) + \hat{z}^l), \qquad \quad l=1, \dots, L
\end{align*}

\subsection{Self-Supervised Joint Pre-training}
The token-based masked language modeling (tMLM) is designed for both speech-text joint pre-training, like RoBERTa ~\cite{roberta}. Given speech or text representations $Z^L$, tMLM predicts the corresponding tokenized tokens $\textbf{c}$ at the masked positions, where each $c_t$ is a $C$-class categorical variable. The distribution over the tokens is parameterized with:
\begin{equation}
    p(c|z^L_t) = \frac{exp(sim(\textbf{W}z^L_t, \textbf{e}(c)) / \tau)}{\sum_{c' \in C} exp(sim(\textbf{W}z^L_t, \textbf{e}(c')) / \tau)}
\end{equation}
where $\textbf{W}$ is a projection matrix, $z_t^L$ is the output hidden state for step $t$ and layer $L$, \textbf{e} is the embedding for token $c$, $sim(a, b)$ computes the cosine similarity between two vectors and $\tau = 0.1$ is used to scale the logits.

Based on the above distribution, we denote the cross-entropy loss computed over masked timesteps as:
\begin{equation}
    \mathcal{L}_{tMLM} = \sum_{t \in \mathcal{M}} log \ p(z_t|z_t^L)
\end{equation}
where $\mathcal{M}$ denotes the set of masked timesteps, and $z_t$ denotes the corresponding token at timestep $t$.

\section{Experiments}
\subsection{Experimental Details}
For unlabeled speech data, we use the publicly available LibriSpeech dataset~\cite{libri}. It contains 960 hours of training data from read audiobooks with three subsets: \textit{train-clean-100, train-clean-360 and train-other-500}, where the labels are not used during pre-training. The unpaired text data are from LibriSpeech LM corpus\footnote{http://www.openslr.org/11/}, containing about 40M English sentences. We randomly select 10\% of the text corpus for our pre-training. For downstream ASR experiments, we use the LibriSpeech \textit{train-clean-100} split, the 1h and 10h splits 
of Libri-Light~\cite{libri-light} as the low-resource labeled data. We use the official baselines for most prior work, which are implemented in fairseq~\cite{fairseq}. Following HuBERT’s best hyper-parameters, the number of kmeans clusters $C$ for speech tokenizer is set to 500 by default unless specified separately. For the lexicon, we use the 200K word-to-phone lexicon provided by LibriSpeech to convert words to phoneme sequences. There are 347 distinct positional-dependent phonemes. 
The network architecture of our token2vec follows that of RoBERT~\cite{roberta}, which is different from HuBERT and wav2vec2.0, since we don't need the convolutional feature extractor for preprocessing waveform input. For token-level MLM, we choose tokens for masking with probability of $0.08$, expanding each chosen token to a span of a length sampled from a normal distribution with mean $10$ and standard deviation $10$. 
We optimize the model with Adam with weight decay $0.01$ and $\beta=(0.9, 0.98)$. The learning rate ramps up linearly for the ﬁrst 32k steps and then is linearly decayed for the following updates. The peak learning rate is 5e-4. All models are trained on 8 GPUs with a batch size of 16384 tokens per GPU for 400k steps. 

For the downstream ASR fine-tuning, we employ 1 hour, 10 hours and 100 hours splits as the supervised data, and use the character set as the model units. We freeze the encoder during the early training, similar to wav2vec2.0~\cite{w2v2}. We use Adam optimizer and a tri-stage schedule where the learning rate is warmed up for the ﬁrst 10\% of updates, held constant for the next 40\% and then linearly decayed for the rest steps. It should be noted that all our results are obtained without language model.

\begin{table}[]
\centering
\begin{tabular}{lcc}
\hline \hline
Model       & test-clean           & test-other           \\ \hline \hline
\multicolumn{3}{l}{\textit{1-hour subset}}  \\ \hline 
wav2vec2.0~\cite{w2v2}    & 24.5     & 29.7    \\
HuBERT~\cite{Hubert}     & 20.9     & 27.5     \\
WavLM~\cite{wavlm}       & 24.5     & 26.7     \\
token2vec(speech-only)        & 22.8     & 28.2     \\
token2vec(speech-text)        & 20.0     & 25.8    \\ \hline \hline
\multicolumn{3}{l}{\textit{10-hour subset}}  \\ \hline 
wav2vec2.0~\cite{w2v2}   & 11.1     & 17.6      \\
HuBERT~\cite{Hubert}     & 10.1     & 16.8      \\
WavLM~\cite{wavlm}       & 9.8      & 16.0      \\
token2vec(speech-only)   & 10.3       & 17.0        \\
token2vec(speech-text)   & 9.0        & 14.9      \\ \hline \hline
\multicolumn{3}{l}{\textit{100-hour subset}}      \\ \hline 
wav2vec2.0~\cite{w2v2}    & 6.1      & 13.3    \\
HuBERT~\cite{Hubert}      & 6.3      & 13.2    \\
WavLM~\cite{wavlm}        & 5.7      & 12.0     \\
token2vec(speech-only)    & 6.2     & 13.1           \\
token2vec(speech-text)    & 5.1     & 11.8    \\ \hline  \hline
\end{tabular}
\caption{WER on LibriSpeech test sets when trained on the LibriLight low-resource labeled data setups of 1 hour, 10 hours and the clean 100h subset of LibriSpeech.}
\vspace{-4mm}
\label{table:main}
\end{table}

\vspace{-2mm}
\subsection{Main Results}
Table~\ref{table:main} presents the ASR results of 1 hour, 10 hours and 100 hours labeled data. The WER is evaluated on the standard LibriSpeech test-clean/other sets. We compare the proposed token2vec with several competitive self-supervised approaches, including wav2vec2.0~\cite{w2v2}, HuBERT~\cite{Hubert} and WavLM~\cite{wavlm}. These three strong baselines are pre-trained on the traditional continuous speech signal. Firstly, from the table, we observe that token2vec pre-trained with only discrete speech tokens can achieve comparable performance (6.2/13.1\%) with wav2vec2.0 and HuBERT, indicating that the discretized speech tokens contain enough semantic information for speech pre-training and downstream speech recognition. These speech tokens are more likely phoneme sequences~\cite{Hubert}, which can be regarded as an intermediary language between raw speech waveform and its transcription.
Secondly, adding additional textual data for joint pre-training, the results show that the proposed token2vec achieves 5.1/11.8\% WER, yielding 17.7/9.2\% relative WER reduction compared with speech-only pre-training, which suggests that additional textual data can benefit the quality of speech representations.

\subsection{Ablation Study}
In this section, we present ablation studies to learn how text token up-sampling and speech token reduction affect the joint pre-training performance in token2vec. We conduct experiments using small models with embedding size 256, four attention heads, and feed-forward embedding size 1,024 in each Transformer block. The pre-training data is 360 hours or 960 hours and fine-tuned on the 100-hour subset.

\vspace{-2mm}
\subsubsection{Effect of Repeating Text Token}
\begin{table}[]
\centering
\begin{tabular}{lcc}
\hline \hline
Model                & test-clean & test-other \\ \hline \hline
\multicolumn{3}{l}{360 hours}                  \\ \hline 
token2vec(speech-only)     & 10.8       & 19.7       \\
token2vec(speech-text-org) & 11.4       & 20.2       \\
token2vec(speech-text)     & 8.6        & 18.2       \\ \hline \hline
\multicolumn{3}{l}{960 hours}                  \\ \hline 
token2vec(speech-only)     & 9.5        & 17.9       \\
token2vec(speech-text-org) & 9.3        & 17.8       \\
token2vec(speech-text)     & 7.3        & 16.2       \\ \hline \hline
\end{tabular}
\caption{Comparison of pre-training with repeated or original text tokens.}
\label{table:text}
\end{table}

Table~\ref{table:text} summarizes the results of joint pre-training with or without repeating text tokens. As shown in the table, joint pre-training with original text sequences gives almost no improvement, suggesting that the length mismatch problem will indeed affect the effectiveness of joint pre-training. 
Moreover, by using only 360 hours of pre-training data, the joint pre-training even surpasses the speech-only pre-training on 960 hours of pre-training data, which indicates that repeated text sequences can be regarded as a kind of speech sequences and help improve the pre-training performance.

\vspace{-2mm}
\subsubsection{Effect of Reducing Speech Token}
\begin{table}[]
\centering
\begin{tabular}{lcc}
\hline \hline
Model                      & test-clean & test-other \\ \hline \hline
token2vec(speech-only, reduced) & 8.5        & 17.2       \\
token2vec(speech-text, reduced) & 8.4        & 17.4       \\
token2vec(speech-only)          & 9.5        & 17.9       \\
token2vec(speech-text)          & 7.3        & 16.2       \\ \hline \hline
\end{tabular}
\caption{Comparision of pre-training with original or reduced speech tokens}
\vspace{-5mm}
\label{table:speech}
\end{table}

Table~\ref{table:speech} reports the results of joint pre-training with or without speech token reduction. Speech-only pre-training with reduced speech tokens is slightly better than that with original speech tokens, however, joint pre-training with reduced speech tokens gives no improvement at all. This might be due to that reducing speech tokens will lose the timing information for pre-training, which is necessary for join pre-training. 

\vspace{-2mm}
\subsection{Transfer to non-ASR Task}
Existing works have shown that self-supervised speech pre-training enjoys good generalizability and reusability across various downstream tasks~\cite{superb, wavlm}. An interesting question is whether our model maintains this advantage if we tokenize the speech into discrete tokens. To answer this question, we evaluate our models on Fluent Speech Commod (FSC) dataset for spoken intent classification (IC), which is from SUPERB benchmark~\cite{superb}. 

Table~\ref{table:fsc} shows the results on downstream IC task. When only pre-training on speech tokens, the proposed token2vec achieves 85.7\% accuracy, better than wav2vec2.0~\cite{w2v2} and DiscreteBERT~\cite{discreteBERT}. When adding additional textual data, token2vec can further improves the accuracy to 96.8\%.
% \begin{table}[]
% \centering
% \begin{tabular}{l|cc}
% \hline \hline
% Model                & IC       & SID      \\ \hline \hline
% FBANK~\cite{superb}  & 9.1      & 8.5E-4    \\
% DiscreteBERT~\cite{discreteBERT}  & 85.7    & 38.8  \\
% wav2vec2.0~\cite{w2v2}            & 92.4    & 75.2  \\
% HuBERT~\cite{Hubert}              & 98.3    & 81.4  \\
% Ours(speech-only)                 & 95.7    &       \\
% Ours(speech-text)                 & 96.8    & 
% \\ \hline \hline
% \end{tabular}
% \caption{Accuracy(\%) on spoken intent classification and speaker identification tasks.}
% \end{table}

\begin{table}[]
\centering
\begin{tabular}{l|c}
\hline \hline
Model                & IC            \\ \hline \hline
FBANK~\cite{superb}  & 9.1     \\
DiscreteBERT~\cite{discreteBERT}  & 85.7    \\
wav2vec2.0~\cite{w2v2}            & 92.4    \\
HuBERT~\cite{Hubert}              & 98.3    \\
token2vec(speech-only)                 & 95.7   \\
token2vec(speech-text)                 & 96.8   \\ \hline \hline
\end{tabular}
\caption{Accuracy(\%) on FSC dataset for spoken intent classification task.}
\vspace{-2mm}
\label{table:fsc}
\end{table}

\subsection{Analysis}
\begin{figure}
    \centering
    \begin{minipage}[t]{0.5\linewidth}
    \centering
    \includegraphics[scale=0.19]{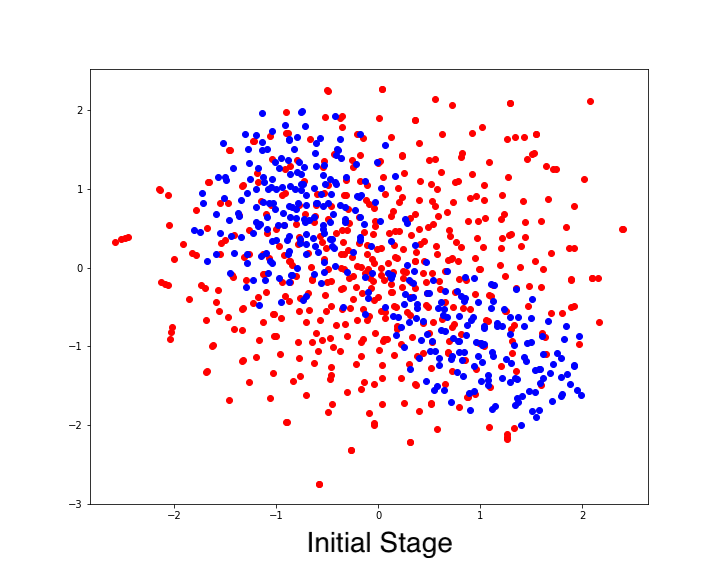}
    \end{minipage}%
    \begin{minipage}[t]{0.5\linewidth}
    \centering
    \includegraphics[scale=0.19]{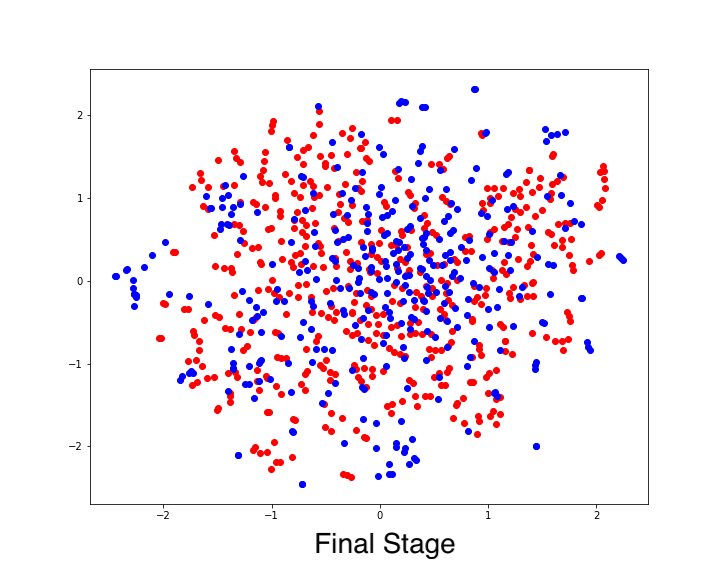}
    \end{minipage}
    \vspace{-2mm}
    \caption{t-SNE plots of speech and text token embeddings at the initial and final stage of the joint pre-training (red points denotes speech and blue point denotes text).}
\vspace{-2mm}
\label{fig:tsne}
\end{figure}

In this section, we want to answer a question: why joint pre-training on unpaired discrete speech and text can improve downstream performance? To answer this question, we use t-SNE to plot the speech and text token embeddings of the initial and final stages of pre-training. As shown in Figure ~\ref{fig:tsne}, as the pre-training goes, speech and text tokens have more and more overlap, suggesting that text sequences can be converted into speech sequences through the embedding layer and vice versa. 

\section{Conclusion}
This paper proposes a novel joint pre-training framework, token2vec, leveraging additional textual data to improve speech pre-training. We design two modality-specific tokenizers for speech and text, i.e. converting continuous speech into discrete speech tokens and words of text into phoneme sequences, respectively. Massive experiments and analysis show that our model significantly outperforms various speech-only pre-training baselines on the LibriSpeech benchmark. Our 
proposed token2vec also shows good transferability on a non-ASR task. For future work, we will investigate the alignment mechanism between speech and text inside the model.

\vfill\pagebreak

\small
\bibliographystyle{IEEEbib}
\bibliography{refs}

\end{document}